\documentclass[10pt,twocolumn,letterpaper]{article}
\usepackage{comment}
\usepackage{makecell}
\usepackage{wacv}              

\definecolor{wacvblue}{rgb}{0.21,0.49,0.74}
\usepackage[pagebackref,breaklinks,colorlinks,allcolors=wacvblue]{hyperref}
\def\wacvPaperID{1607} 
\def\confName{WACV}
\def\confYear{2026}
\usepackage{soul}
\sethlcolor{yellow}

\title{RobuMTL: Enhancing Multi-Task Learning Robustness \\ Against Weather Conditions}

\author{Tasneem Shaffee\\
Brown University\\
Providence, RI\\
{\tt\small tasneem\_shaffee@brown.edu}
\and
Sherief Reda\\
Brown University\\
Providence, RI\\
{\tt\small sherief\_reda@brown.edu}
}
\begin{document}
\maketitle
\begin{abstract}
 Robust Multi-Task Learning (MTL) is crucial for autonomous systems operating in real-world environments, where adverse weather conditions can severely degrade model performance and reliability. In this paper, we introduce RobuMTL, a novel architecture designed to adaptively address visual degradation by dynamically selecting task-specific hierarchical Low-Rank Adaptation (LoRA) modules and a LoRA expert squad based on input perturbations in a mixture-of-experts fashion. Our framework enables adaptive specialization based on input characteristics, improving robustness across diverse real-world conditions. To validate our approach, we evaluated it on the PASCAL and NYUD-v2 datasets and compared it against single-task models, standard MTL baselines, and state-of-the-art methods. 
 On the PASCAL benchmark, RobuMTL delivers a +2.8\% average relative improvement under single perturbations and up to +44.4\% under mixed weather conditions compared to the MTL baseline. On NYUD-v2, RobuMTL achieves a +9.7\% average relative improvement across tasks. 
The code is available at GitHub 
{\footnote{\url{https://github.com/scale-lab/RobuMTL.git}}}.

 

\end{abstract}    
\section{Introduction}
\label{sec:intro}
While extensive research has been conducted on single-task learning, many real-world applications, such as virtual reality systems and autonomous vehicles, require the simultaneous execution of multiple tasks in an computing environment with constrained resource, such as object segmentation and depth estimation \cite{DBLP:conf/aaai/GhamiziCPT22}. Multi-Task Learning (MTL) emerges as a powerful training paradigm designed to address such needs by combining data from multiple tasks and training them concurrently \cite{Crawshaw2020MultiTaskLW}. MTL leverages shared representations to capture common patterns across tasks while tailoring the learning process to optimize each task for its specific objectives \cite{Crawshaw2020MultiTaskLW}. This approach often results in superior performance, owing to its enhanced generalization capabilities derived from learning across diverse tasks.

Robustness is critical for multi-task learning (MTL) in real-world applications such as autonomous driving, where distribution shifts from adverse weather or sensor noise can severely degrade performance. While robustness has been widely explored in single-task settings, for example, semantic segmentation \cite{kerim2022Semantic, gella2023weatherproof, 9294554} and object detection \cite{10483822,10484404, Jeon2023DARAWDA, CHEN2025128994, chu2024dyolorobustframeworkobject}, MTL robustness remains underexplored. 

Existing approaches such as adversarial training or data augmentation often improve robustness at the cost of clean accuracy, and this trade-off becomes more challenging when multiple tasks share features that are affected unevenly by perturbations, amplifying task conflicts \cite{Kalb2023PrinciplesOF}. To our knowledge, this is the first work investigating robustness for MTL under adverse weather conditions.

Pre-processing methods such as restoration models \cite{9157460, chen2020pre, kawar2022jpeg, valanarasu2021transweather, 10.1007/978-3-031-19797-0_26, AirNet} can improve degraded inputs but increase latency and memory, making them less suitable for real-time MTL and offering no consistent task-wise benefits. Mixture-of-Experts (MoE)-based MTL methods dynamically route inputs to experts, yet routers often misclassify noisy patterns and incur instability and overhead by being invoked at every layer.

Finally, most MTL models rely on static inference paths, ignoring input conditions during deployment. This rigidity prevents dynamic adaptation to varying input qualities. In particular, mixed perturbations, where clean and perturbed inputs coexist are poorly handled by existing methods, as training jointly on such diverse inputs further complicates optimization and compromises overall task performance.
In this paper, we aim to address the trade-off between robustness and preserving model quality, particularly in MTL, while also mitigating the mentioned issues. 
Our contributions can be summarized as follows.
\begin{itemize}
    \item We propose RobuMTL, a pipeline that dynamically selects and applies lightweight hierarchical LoRA experts to encoder based on input characteristics, mitigating weather-induced degradation through a MoE-inspired expert aggregation strategy.
    \item We extend this approach (RobuMTL+) by incorporating dynamically selected modular adaptations, where perturbation-specific hierarchical LoRA experts are complemented with task adaptive expert modules per perturbation.
    \item We introduce a Dynamic Modular LoRA Selector (DMLS) that dynamically assigns importance scores to LoRA and parameter experts during aggregation, reflecting their contribution with respect to the input based on perturbation type.

    
    
    \item 
    We performed an analysis to evaluate the performance of traditional MTL and single-task  against different adverse weather condition on two datasets. Additionally, we compared our approach with state-of-the-art (SOTA) models, demonstrating its superiority over other methods.
  

\end{itemize}

The paper is organized as follows. Section \ref{relatedwork} discusses the related work. Section \ref{method} illustrates the methodology. Section \ref{results} demonstrates the experimental setup, dataset generation and shows the obtained results. Section \ref{conclusion} concludes the work and illustrates the future work.

\section{Related Work} \label{relatedwork}
\textbf{MTL:} Prior work has explored improving feature sharing and task interaction in MTL through attention, distillation, and parameter-efficient adaptation \cite{MRK19, 10.1007/978-3-030-58548-8_31, padnet, Agiza2024MTLoRAAL, hu2021loralowrankadaptationlarge, taskprompter2023}.

\textbf{Robustness to Adverse Weather:} Robust perception under weather-induced distribution shifts has been studied primarily in single-task or limited MTL scenarios, focusing on weather-aware adaptation or joint classification/segmentation \cite{10.1145/3653781.3653810,8641680,XIE2022116689}.

\textbf{Mixture-of-Experts (MoE):}
The concept of Mixture of Experts (MoE) involves combining multiple submodels, or experts, and dynamically selecting which ones to activate based on the input, enabling efficient conditional computation. 

Zhang \textit{et al.} proposed UIR-LoRA, which leverages a generative backbone and integrates pretrained LoRAs at the output stage in an MoE-like manner \cite{zhang2024uirloraachievinguniversalimage}.
Chen \textit{et al.} (Mod-Squad) and Liang \textit{et al.} (M3ViT) incorporated MoE into vision transformers, activating parameter groups or sparse backbone experts via routing \cite{chen2022modsquaddesigningmixtureexperts, liang2022m3vitmixtureofexpertsvisiontransformer}.
More recent works extend LoRA into MoE: MeteoRA \cite{xu2025meteora}, MLoRE \cite{jiang2024mlore}, and MixLoRA \cite{li2024mixloraenhancinglargelanguage} design hybrid or fusion-style expert structures; MOELoRA \cite{liu2023moelora} applies task-wise Top-1 routing for medical MTL; MoCLE \cite{gou2024mixtureclusterconditionalloraexperts} and MoRAL \cite{yang2024moralmoeaugmentedlora} adopt token-wise Top-1/Top-K routing for multi-modal or lifelong learning; MoELoRA \cite{liu2024moe} and MOLA \cite{gao2024higher} explore Top-K token routing with and without contrastive balancing.

In contrast, RobuMTL introduces hierarchical rank LoRA plugins across attention and FFN layers but invokes the router only once before the main MTL backbone. This avoids the training instability of repeated MoE routing while directly targeting robustness under perturbations.

\section{{Methodology}} \label{method}
\begin{figure*}[t]
\centering
\captionsetup{justification=justified,singlelinecheck=false}
\includegraphics[width=\textwidth]{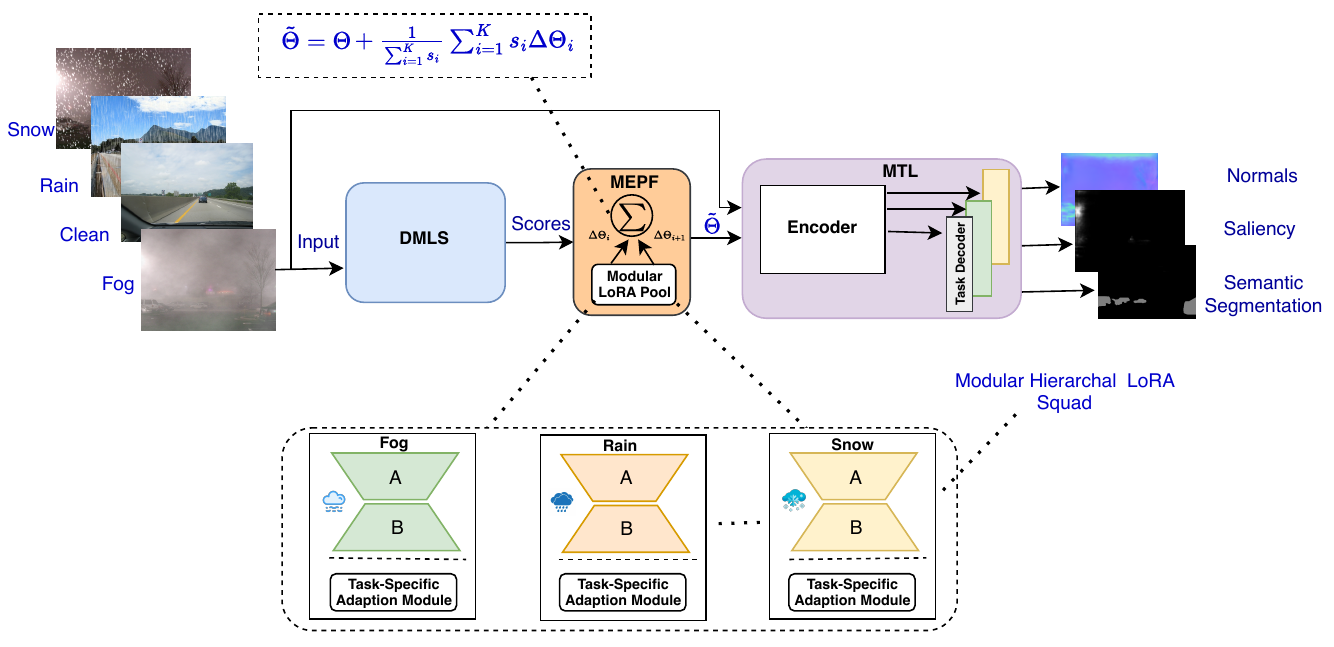}
\caption{{
In our pipeline, the input image is first processed by DMLS to estimate contribution scores based on its characteristics, and MEPF then uses these scores to selectively aggregate the relevant LoRA experts and update the MTL weights.The number of aggregated experts is at most equal to the number of perturbations.
}}
\label{fig:design}
\end{figure*}
Our methodology comprises three components: (1) a multi-task learning backbone for four tasks, (2) the Dynamic Modular LoRA Selector (DMLS), and (3) the Multiple Expert Parameter Fusion (MEPF) module as shown in Figure \ref{fig:design}. The architecture uses a hierarchical MTL design with a shared Swin-ViT encoder and lightweight task-specific decoders based on linear and bilinear upsampling. Following Agiza \textit{et al.} \cite{Agiza2024MTLoRAAL}, LoRA weights are inserted into attention, projection, and feed-forward layers. Unlike their uniform configuration, we introduce a hierarchical rank allocation: lower ranks in early layers and higher ranks in deeper layers. This inverted-triangle design preserves parameter efficiency while assigning adaptation capacity where semantics are learned. 
It also reduces overfitting to low level degradations in early visual features while strengthening robustness and multitask representation learning in later stages with minimal overhead.
\subsection{MTL Preliminaries}
\textbf{MTL:} We utilize a pre-trained MTL model composed of a shared encoder and task-specific decoders. Given a dataset $\mathcal{D}_t = \{(x_i, y_i^t)\}_{i=1}^{N_t}$, where $N_t$ is total number of samples for task $t$, the shared encoder with parameters {$\Theta$} processes input $x_i$ to produce an intermediate shared representation:
\begin{equation}
h_i = f_{\text{shared}}(x_i; \Theta).
\end{equation}

Each task-specific decoder with parameters $\Theta_t$ predicts the outputs using this shared representation ($h_i$) as input:
\begin{equation}
\hat{y}_i^t = f_t(h_i; \Theta_t), \quad \text{for } t = 1, \dots, T,
\end{equation}
where T is total number of the tasks. The total MTL loss combines all task-specific losses, weighted by $\lambda_t$ to balance task importance:
\begin{equation}
\mathcal{L}_{\text{MTL}} = \sum_{t=1}^T \lambda_t \mathcal{L}_t(\hat{y}_i^t, y_i^t),
\end{equation}
where $\hat{y}_i^t,$ is the task output at given sample $i$, $y_i^t$ is the target. 
The task-specific loss is denoted as $\mathcal{L}_t$, which is defined according to the nature of each task $t$. 
The training objective is to jointly optimize both shared encoder parameters $\Theta$ and all task-specific decoder parameters $\{\Theta_t\}_{t=1}^T$.

\textbf{MTL with LoRA:}
LoRAs are small decomposed matrices $ \mathbf{A}$ and $ \mathbf{B}$ that are adapted for complex pretrained models, where the low-rank update is $\boldsymbol{\Delta\Theta}=\mathbf{AB}$. They are added to the weights at the end, where the output layer with LoRA is given  as follows.
\begin{equation}
\boldsymbol{\tilde{\Theta}} = {(\boldsymbol{\Theta} + \alpha \cdot \mathbf{AB}) \mathbf{x},}
\end{equation}
where  $\boldsymbol{\Theta}$ is the frozen pretrained weight matrix, $\alpha$ is the adaptation scale that regulates the extent of deviation of the tuned model from the original model, and $\boldsymbol{\tilde{\Theta}}$ represents the effective weight matrix after incorporating the LoRA update.
LoRA adaptations are introduced following a hierarchical task-aware scheme within the MTL architecture. Rather than uniformly inserting task-agnostic and task-specific LoRA modules across all layers, we selectively apply hierarchical task-specific LoRA only within the final stage of the Swin Transformer. This placement focuses adaptation where feature representations are most abstract and task-relevant. LoRA ranks are configured hierarchically across this stage, allowing lower layers within the stage to emphasize shared features with smaller ranks, while deeper layers prioritize task-specific refinements through higher-rank adaptations. This structured distribution balances generalization and specialization, ensuring effective multi-task learning without overwhelming earlier model layers.

\subsection{{ Dynamic Modular LoRA Selector (DMLS)}}

\begin{figure*}[t]

\includegraphics[width=\textwidth]{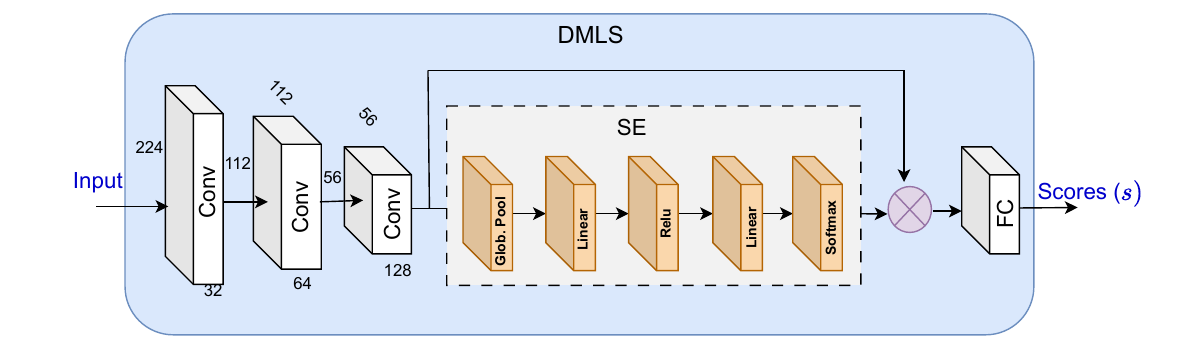}
\caption{{
DMLS: Dynamic Modular LoRA Selector.
}}
\label{fig:dmls}
\end{figure*}

The DMLS module serves as a lightweight policy network that selects which pretrained LoRAs with different ranks and task related parameters should be combined with the base model during inference. Unlike previous weather classification methods that rely on heavy CNN-RNN or VGG based architectures \cite{ZHAO201847, 7351424}, our joint weather and degradation classifier contains only 48,742 parameters. DMLS is based on CNN that predicts input conditions and maps them to the appropriate pool of LoRAs or modular of parameters. It operates directly on the input image and is applied once before the MTL model, reducing routing overhead and instability commonly observed in MoE based expert selection.
The architecture of the network consists of three main components: 1) a feature extraction backbone, 2) a squeeze-and-excitation (SE) block \cite{8578843} for channel attention, and 3) a fully connected classifier (FC) as shown in Figure \ref{fig:dmls}. 
Below, we describe each component in detail. 

\textbf{1. Feature Extraction Backbone:} 
The classifier uses a CNN backbone to extract hierarchical local features needed to distinguish different perturbations such as noise, blur, rain, and snow. It includes three convolutional stages with max pooling for downsampling.
Depthwise separable convolutions are used to reduce computation, expanding features to 128 channels in the final stage, followed by adaptive average pooling for global downsampling.

\begin{figure}[h!]
  \centering
  \includegraphics[width=\columnwidth]{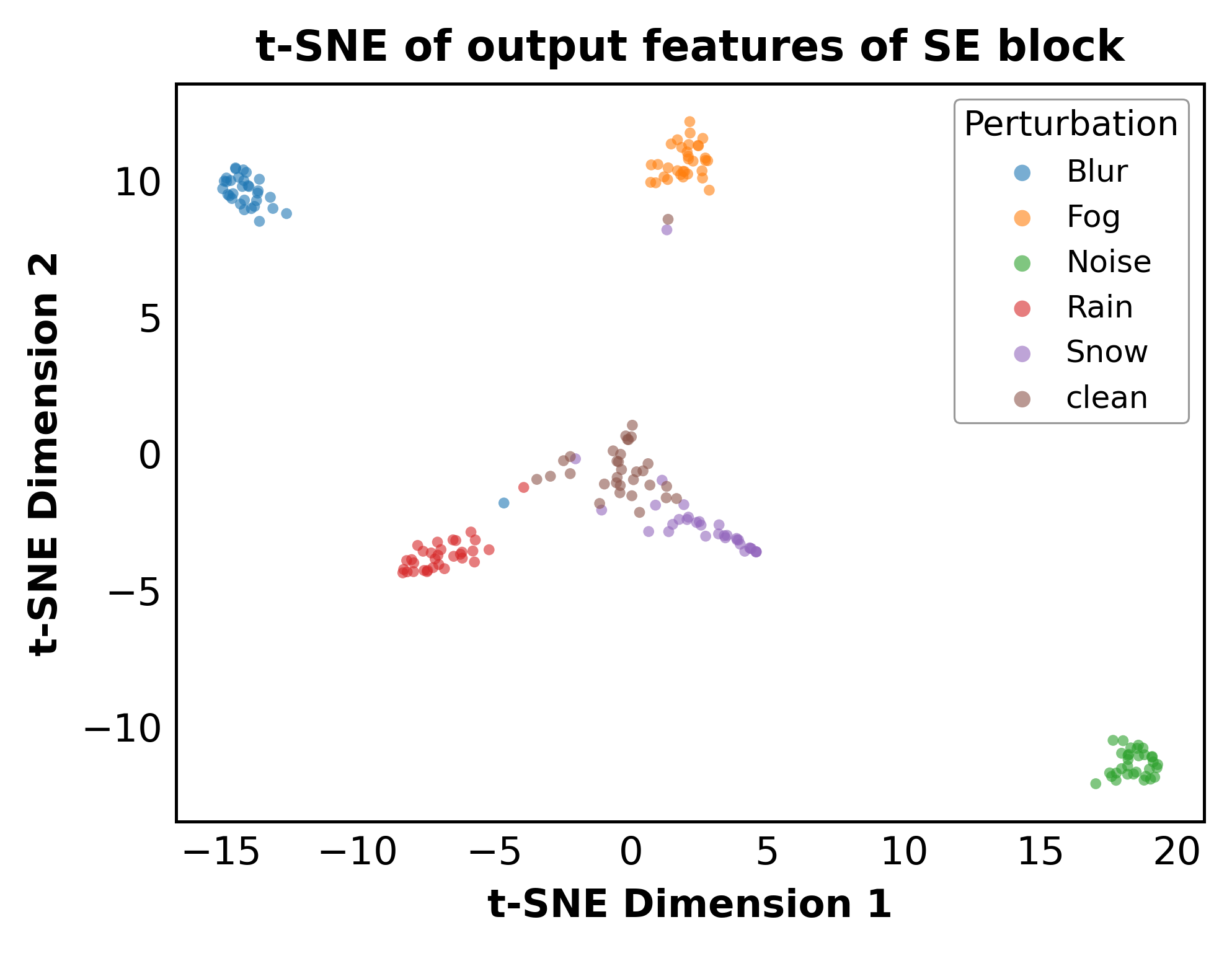}
   \caption{Illustration of SE contribution within DMLS.}
   \label{fig:tsne}
\end{figure}

\textbf{2. Squeeze-and-Excitation (SE) Block:}
SE block \cite{8578843} is applied to the output of the backbone to enhance the model’s representational power by adaptively recalibrating channel-wise feature responses. 
The block first squeezes spatial dimensions into a global feature descriptor, which provides a compact summary of the feature map. The Exciting part is to obtain the important feature channels by assigning higher attention weights based on their relevance to the input. 
The SE block computes channel-wise attention weights in our case as follows:
\begin{align}
\mathbf{z} &= \text{ReLU}(\mathbf{W}_1 \cdot \mathbf{F}_\text{global}), \\
\mathbf{e} &= \text{Softmax}(\mathbf{W}_2 \cdot \mathbf{z}),
\end{align}
where $\mathbf{W}_1 \in \mathbb{R}^{128 \times 64}$ and $\mathbf{W}_2 \in \mathbb{R}^{64 \times 128}$ are fully connected layers. $\mathbf{F}_\text{global}$ is obtained by applying global average pooling to the backbone feature maps, and $\mathbf{e}$ is the output of SE block.
We replaced the sigmoid layer in conventional SE with softmax, as it performs better by emphasizing dominant features while suppressing irrelevant ones.
The resulting attention weights $\mathbf{e}$ are applied to the features:
\begin{equation}
\mathbf{F}_\text{weighted} = \mathbf{F}_\text{global} \odot \mathbf{e},
\end{equation}
where $\odot$ denotes element-wise multiplication.
The capability of the SE block to distinguish between different perturbation types is illustrated in Figure \ref{fig:tsne}, where a t-SNE projection shows clear separation in the feature space.

\textbf{3. Fully Connected Classifier:}
The final classification head consists of two fully connected layers.
The final scores are generated by applying softmax over the output layer of fully connected classifier as follows.
\begin{align}
\mathbf{s} &= \text{Softmax}(\mathbf{{F}_\text{weighted}}),
\end{align}
where $\mathbf{s}$ denotes the normalized scores (i.e., probability distribution) over $n$ classes, with $n$ being the number of perturbations. 
The DMLS is trained separately on all perturbations using a supervised learning strategy, where synthetic perturbations are generated and labeled accordingly. Unlike traditional routers in mixture-of-experts approaches, its loss is computed directly against the ground-truth perturbation types.
The cross entropy loss function is used for the DMLS which is defined as follows.
\begin{equation}
\mathcal{L}_{\text{selector}} = - \frac{1}{N} \sum_{i=1}^{N} \log s_{i, y_i},
\end{equation}
where ${N}$ is the number of training samples, ${\log {s}_{i, y_i}}$ is the log-probability of the ground-truth class ${y_i}$ under the model’s predicted distribution ${s_i}$.
\subsection{Multiple Expert Parameter Fusion (MEPF)}
We found that naively integrating the outputs of multiple hierarchal LoRAs (e.g., via simple averaging) significantly degrades performance. 
{To address this, we introduce the MEPF module, which adaptively integrates the selected hierarchical LoRAs based on their average contribution scores from the Top-K selection. 

The DMLS generates scores for all $n$ perturbations, from which $k$ LoRA experts are selected using MEPF.
This enables our system to handle composite perturbations (e.g., rain combined with fog), where a single LoRA may be insufficient for mitigation.

The MEPF maintains a pool of LoRA experts $ \mathbf{\Delta  \Theta_{i}, \Delta  \Theta_{i+1}, \Delta  \Theta_{K}}$ and can be extended to incorporate a modular pool of hierarchal LoRAs or other task-related parameters, such as decoder headers. Each specialized squad consists of a perturbation-specific hierarchal LoRA for the backbone, along with its corresponding normalization and decoder head parameters, ensuring adaptability to diverse perturbation types. 
Top-K selection determines how many experts are combined during inference. We find that $k{=}1$ works best for single perturbations, while $k{=}6$ is optimal for mixed conditions.
Here, $k$ denotes the number of experts combined via weighted averaging to mitigate perturbations, with final parameters updated as: 
\begin{equation}
    \boldsymbol{\tilde{\Theta}} = \boldsymbol{\Theta} + 
    \left(\frac{1}{\sum_{i=1}^{K} s_i}\right) 
    \sum_{i=1}^{K} s_i \, \boldsymbol{\Delta \Theta_i},
\end{equation}
 where $\sum_{i=1}^{K} s_i \mathbf{\Delta \Theta_i}$ represents the total weighted aggregation of the LoRA experts ($\mathbf{\Delta \Theta_i}$). Here, $s_i$ is the score generated by the DMLS, indicating the contribution weight of each perturbation in the image. The term $\sum_{i=1}^{K} s_i$ denotes the sum of the scores of the selected LoRA experts (i.e., the total weight), which serves as a normalization factor when the number of selected experts is smaller than the total number of perturbations, since the DMLS always produces scores for all perturbations. This value approaches $1$ when all experts are aggregated. 
 When applying a single hierarchical LoRA or hierarchical LoRA squad at a time, we select the top-1 option as the best case (highest scores) as follows. 
\begin{equation}
\text{TopK}(s, 1)_i =
\begin{cases} 
s_i, & \text{if } s_i = \max_j s_j, \\
0, & \text{otherwise}.
\end{cases}
\end{equation}
\subsection{{Training and Inference}}
We have observed that the conventional MoE training setup becomes unstable with corrupted inputs, leading to slow learning and suboptimal expert performance. Therefore, we adopted a technique that deactivates all non-relevant experts during training, ensuring only the required expert is updated and avoiding conflicts.
Our first training strategy, RobuMTL, trains one hierarchical LoRA expert per perturbation type (rain, snow, noise), while using a shared decoder. We warm up on clean data for stable convergence, then train on the target perturbation while mixing a small portion of other data to reduce overfitting. The clean expert is trained only on clean samples. We also observe that training solely on noise or blur harms the normals task, whereas broader data exposure improves generalization. Finally, we apply a short fine tuning step to adapt the shared decoder and normalization layers to the combined expert distribution.
The second strategy (RobuMTL+) introduces task-specific adaptation modules that accompany each hierarchical LoRA expert (forming a hierarchical expert squad). In this design, we train one LoRA expert together with its paired adaptation modules at a time, allowing them to jointly specialize for a specific perturbation.

During inference, the DMLS predicts the most suitable perturbation-specific hierarchical LoRAs or LoRA squads based on the input image, which are then aggregated by MEPF. The selected experts are applied to the backbone before processing, enabling dynamic adaptation to conditions such as snow or noise.

\section{Experimental Results} \label{results}
Experiments are conducted using a Swin-Tiny backbone trained in PyTorch on an NVIDIA V100 GPU. 
We used two datasets for evaluation. The DMLS is trained on one dataset and successfully generalizes to another.

\begin{table*}[t]
\centering
\caption{
Overall performance comparison of our proposed model against baselines and state-of-the-art methods on unseen clean and adverse weather conditions. Bold indicates the best performance.
}
\label{table:results}
\setlength{\tabcolsep}{3pt}
\begin{tabular}{lrrrrrrrrr}
\toprule
\multicolumn{1}{l}{\textbf{Method}} &
\multicolumn{1}{c}{\textbf{\makecell{SemSeg\\(mIoU) $\uparrow$}}} &
\multicolumn{1}{c}{\textbf{\makecell{Normals\\(RMSE) $\downarrow$}}} &
\multicolumn{1}{c}{\textbf{\makecell{Saliency\\(mIoU) $\uparrow$}}} &
\multicolumn{1}{c}{\textbf{\makecell{HP\\(mIoU) $\uparrow$}}} &
\multicolumn{1}{c}{\textbf{\makecell{$\Delta m_{\text{adv}}$\\(\%)}}} &
\multicolumn{1}{c}{\textbf{\makecell{$\Delta m_{\text{clean}}$\\(\%)}}} &
\multicolumn{1}{c}{\textbf{\makecell{Params\\(M)}}} &
\multicolumn{1}{c}{\textbf{\makecell{FLOPs\\(G)}}} &
\multicolumn{1}{c}{\textbf{FPS}} \\
\midrule
Single Task            & 61.61 & 18.97 & 61.97 & 53.78 &   0.00 & +10.00 & 112.60 &  73.70 & 65 \\
MTL-Full FT            & 58.56 & 19.35 & 61.40 & 52.33 &  -2.60 &  +7.70 &  29.50 &  21.47 & 84 \\
MTL-Tuning Decoders    & 53.17 & 21.96 & 54.70 & 46.56 & -13.30 &  -3.80 &  29.50 &  21.47 & 84 \\
MTLMoE                 & 39.10 & 20.90 & 56.60 & 41.30 & -24.00 &  -9.70 & 115.90 &  77.16 & 22 \\
\midrule
Clean Single Task               & 52.90 & 20.69 & 58.09 & 48.45 &  -9.80 &   0.00 & 112.60 &  73.70 & 65 \\
Clean MTL-Full FT      & 52.59 & 20.97 & 56.02 & 47.59 & -11.50 &  -1.80 &  29.50 &  21.47 & 84 \\
Clean MTL-Decoders     & 47.23 & 24.70 & 48.59 & 41.07 & -24.70 & -15.40 &  29.50 &  21.47 & 84 \\
\midrule
Clean MTLoRA-r16       & 50.35 & 20.95 & 55.50 & 46.86 & -13.00 &  -3.50 &  30.89 &  20.88 & 43 \\
Clean MTLoRA-r32       & 49.56 & 21.10 & 55.66 & 47.27 & -13.30 &  -3.70 &  32.02 &  20.88 & 41 \\
Clean MTLoRA-r64       & 48.05 & 20.55 & 56.43 & 46.95 & -13.00 &  -3.60 &  34.26 &  20.88 & 37 \\
\midrule
MTI-Net (small)        & 19.50 & \textemdash{}  & 43.34 & 28.83 & -48.30 & -43.00 &  12.64 &  19.10 & 12 \\
PAD-Net                & 13.29 & 36.09 & 42.72 & 29.07 & -61.40 & -53.90 &  19.51 & 265.00 & 11 \\
ASTMT                  & 52.48 & 21.73 & 61.30 & 52.14 &  -8.40 &  +1.80 &  31.27 &  25.91 & 15 \\
Meteora (SwinT)        & 39.10 & 20.70 & 58.76 & 41.60 & -18.50 &  -9.70 &  98.57 &  60.44 & 21 \\
\midrule
MTLoRA (r=16)          & 57.54 & 19.61 & 60.34 & 50.89 &  -4.50 &  +5.70 &  30.89 &  20.88 & 43 \\
MTLoRA (r=32)          & 53.97 & 20.40 & 57.70 & 49.11 &  -8.90 &  +1.00 &  32.00 &  20.88 & 41 \\
MTLoRA (r=64)          & 52.70 & 20.20 & 57.66 & 52.17 &  -7.80 &  +2.20 &  34.26 &  20.88 & 37 \\
\midrule
\textbf{RobuMTL}       & \textbf{63.11} & \textbf{18.52} & \textbf{62.75} & \textbf{53.81} & \textbf{+1.53} & \textbf{+12.21} & 30.92 & 21.50 & 59 \\
\textbf{RobuMTL+}      & \textbf{63.86} & \textbf{18.17} & \textbf{62.64} & \textbf{55.01} & \textbf{+2.80} & \textbf{+13.50} & 30.92 & 21.50 & 57 \\
\bottomrule
\end{tabular}
\end{table*}

\textbf{Dataset Generation:}
The PASCAL  \cite{everingham2010voc} and NYUDv2 \cite{{Silberman2012IndoorSA}} datasets were selected due to its widespread use and recognition as benchmark datasets for MTL. 
To simulate adverse weather changes and image corruption, we created a replica of the PASCAL and NYUD datasets and generated five additional datasets alongside the original. Each dataset contains a specific type of perturbation and is used to train and evaluate the LoRAs, DMLS, and SOTA models.

 \textit{Noise, Blur, and Natural Perturbations}: Environmental factors like rain, snow, fog, noise, and blur can degrade visual input quality. To assess model robustness, we created multiple variations of the original dataset, each featuring a specific corruption type. Gaussian noise was added at varying severity levels, while blur corruptions included average, motion, and Gaussian blurring. Rain and snow datasets were modified by adjusting drop and snowflake sizes, respectively. The datasets were generated using the \texttt{imgaug} library \footnote{\url{https://imgaug.readthedocs.io/en/latest/source/api_augmenters_weather.html}}, which provides various augmentation techniques for simulating environmental conditions. 

  
 
    
    

\textbf{Metric Evaluation:}
The evaluation metrics for the tasks are as follows: semantic segmentation, saliency estimation, and human part segmentation are assessed using mean Intersection over Union (mIoU). For the surface normals task and depth estimation, performance is measured using the root mean square error (RMSE) of the predicted angles. Edge detection is assessed by reporting its loss. Additionally, the overall performance, denoted as $\Delta m$, is calculated as the average reduction in performance per task relative to the single-task baseline (st).
\begin{equation}
   \Delta m = \frac{1}{T} \sum_{i=1}^T (-1)^{l_i} \frac{M_i - M_{st,i}}{M_{st,i}},
\end{equation}
where ${l_i=1}$ if a lower value indicates improved performance for the metric ${M_i}$ of task ${i}$ and ${l_i=0}$ otherwise. ${M_{st,i}}$ denotes the performance metric of the corresponding single task $i$. The single-task performance is determined using a fully trained model designed specifically for that task, utilizing the same backbone network. In our case, we measure ${\Delta m_\text{adverse}}$ and ${\Delta m_\text{clean}}$. The term ${\Delta m_\text{adverse}}$ quantifies the relative overall performance of the model against an adverse single-task model (trained on an adverse weather condition datasets), assessing how much the model degrades compared to this baseline. Meanwhile, ${\Delta m_\text{clean}}$ captures the change in accuracy relative to a clean single-task model (trained only on clean data), evaluating how well the model's accuracy aligns with or deviates from the clean single-task baseline.


\textbf{Qualitative Analysis:
}
For the approach of modifying the inference pipeline, we assume that the model (UIR-LoRA) \cite{zhang2024uirloraachievinguniversalimage}, referenced in Section \ref{relatedwork}, with a total of $95.2$ (M) parameters, performs optimally and generates high-resolution restored images using a generative model with LoRAs. However, incorporating this approach significantly increases the overall model complexity to $125.256$ (M) parameters, including the total MTL parameters of $30$ (M). Therefore, our work focuses on leveraging a parameter-efficient training approach for MTL to enhance robustness while minimizing inference complexity. We specifically aim to compare our method against the conventional approach of training and fine-tuning pretrained models on adverse weather data.

We compared the performance of our proposed framework against the baseline 1) fully tuned single task, 2) traditional full fine-tuned MTL, and 3) traditional MTL with fine-tuned decoders only, trained on clean and adverse datasets, 4) conventional MTLMoE, and 5) SOTA models, such as MTLoRA \cite{Agiza2024MTLoRAAL}, ASTMT \cite{MRK19}, MTI-Net \cite{10.1007/978-3-030-58548-8_31}, MLoRE \cite{jiang2024mlore}, Meteora \cite{xu2024meteoramultipletasksembeddedlora}, TaskPrompter \cite{taskprompter2023}, PAD-Net \cite{padnet}. Our approach, along with traditional MTL, single-task models, and MTLoRA, utilizes a Swin-Tiny backbone, whereas PAD-Net and MTI-Net are based on HRNet-18, and ASTMT relies on a ResNet26 backbone, MLoRE and Taskprompter are based on Vit. A fine-tuning training process is performed on the models across six datasets to generate adversarially trained models. We also compared our approach with several methods designed to address task conflicts, such as NashMTL \cite{navon2022multi}, PCGrad \cite{yu2020gradient}, and CAGrad \cite{liu2021conflict}, applied to MTL (Swin-Tiny). In our comparison, we evaluate RobuMTL, which integrates LoRAs and undergoes fine-tuning after integration. In contrast, RobuMTL+ incorporates both LoRAs and their modular squad without any additional fine-tuning.

From Tables  \ref{table:results}, \ref{tab:pascal_}, and \ref{tab:nyud}, the performance of SOTA models has degraded as it forces the model to divide its capacity between learning both clean and adversarial examples. This trade-off limits the model's ability to fully optimize for either clean or adverse data, ultimately reducing its overall effectiveness. RobuMTL(+) outperforms both the baselines and state-of-the-art methods on PASCAL, benefiting from its expert modules specialized for each perturbation. RobuMTL+ further achieves the highest performance among all methods on NYUD-v2. On clean data, clean RobuMTL(+) also improves accuracy by +3.18\% over the clean single-task baseline. 
We use the (16,32,64,128) LoRA configuration for PASCAL and a hierarchical rank starting at 64 (64,128,256,512) for NYUD-v2, balancing capacity to recover corrupted structures while avoiding noise amplification, especially in edge and depth tasks.
The  LoRA squad members are aggregated on the CPU, and only the aggregated squad is transferred to the GPU during inference. 
As shown in Table \ref{table:results}, RobuMTL achieves a 3.64× reduction in parameters and 3.43× lower computation cost over single-task training.
FPS in Table \ref{table:results} is measured on an NVIDIA V100 GPU (batch size = 1). 
RobuMTL(+) slightly reduces FPS due to LoRA routing but remains faster than most of existing robust MTL baselines, retaining practical deployability.
Figure \ref{fig:experts} shows the expert activations for each perturbation on NYUD-v2, demonstrating the stability of DMLS to effectively select experts without being retrained on NYUD-v2.
\begin{figure}[t]
  \centering
  \includegraphics[width=\columnwidth]{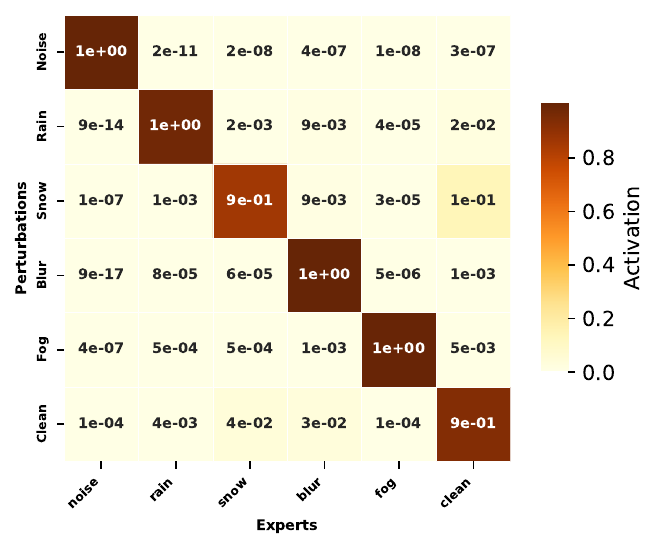}
   \caption{
   DMLS expert activations under single perturbations show diagonal dominance(high activation) for the matching expert and minimal activation for others.
   }
   \label{fig:experts}
\end{figure}
We also evaluate our approach on Swin-Base for a fair comparison with TaskPrompter and MLoRE, as shown in Table~\ref{tab:pascal_}. 


\begin{table}[t]
\centering
\caption{
PASCAL performance comparison using Swin-B  and ViT-B backbones.
Bold denotes best performance.
}
\label{tab:pascal_}
\resizebox{\columnwidth}{!}{%
\setlength{\tabcolsep}{2.5pt}
\begin{tabular}{lrrrrrrr}
\toprule
\textbf{Model} &
\textbf{\makecell{Seg\\(mIoU) $\uparrow$}} &
\textbf{\makecell{Norm\\(RMSE) $\downarrow$}} &
\textbf{\makecell{Sal\\(maxF) $\uparrow$}} &
\textbf{\makecell{HP\\(mIoU) $\uparrow$}} &
\textbf{\makecell{$\Delta m$\\(\%)}} &
\textbf{\makecell{FLOPs\\(G)}} &
\textbf{\makecell{Params\\(M)}} \\
\midrule
Single Task     & 69.50 & 18.10 & 75.80 & 57.30 &  0.00 &  61.64 & 349.20 \\
TaskPrompter    & 69.00 & \textbf{17.10} & 80.30 & 54.30 & +1.40 & 1134.08 & 418.33 \\
MLoRE           & 67.60 & 17.30 & \textbf{81.00} & 55.60 & +1.40 & 890.90 & 259.13 \\
\midrule
\textbf{RobuMTL}  & \textbf{71.50} & 18.00 & 77.30 & \textbf{62.20} & \textbf{+3.50} &  63.50 & 101.62 \\
\textbf{RobuMTL+} & \textbf{72.35} & 17.50 & 78.00 & \textbf{63.10} & \textbf{+5.00} &  63.50 & 101.62 \\
\bottomrule
\end{tabular}}
\end{table}

\begin{table}[t]
\centering
\caption{NYUD-v2 performance. 
Swin-Tiny is used as the backbone for all models, except PAD-Net, which uses HRNet-18.
}
\label{tab:nyud}
\resizebox{\columnwidth}{!}{%
\setlength{\tabcolsep}{2.5pt}
\begin{tabular}{lrrrrrrr}
\toprule
\textbf{Model} &
\multicolumn{1}{c}{\textbf{\makecell{Seg\\(mIoU) $\uparrow$}}} &
\multicolumn{1}{c}{\textbf{\makecell{Norm\\(RMSE) $\downarrow$}}} &
\multicolumn{1}{c}{\textbf{\makecell{Edge\\(Err) $\downarrow$}}} &
\multicolumn{1}{c}{\textbf{\makecell{Depth\\(RMSE) $\downarrow$}}} &
\multicolumn{1}{c}{\textbf{\makecell{$\Delta m$\\(\%)}}} &
\multicolumn{1}{c}{\textbf{\makecell{FLOPs\\(G)}}} &
\multicolumn{1}{c}{\textbf{\makecell{Params\\(M)}}} \\
\midrule
Single Task          & 33.25 & 29.84 & 0.08 & 0.72 &   0.00 &  73.70 & 112.60 \\
MTLMoE      & 17.91 & 34.63 & 0.08 & 0.91 & -22.10 &  77.16 & 115.90 \\
MTLoRA      & 30.80 & 30.27 & 0.08 & 0.75 &  -3.20 &  20.80 &  34.26 \\
PAD-Net     & 33.92 & \textbf{21.90} & 0.08 & 0.67 &  +9.20 & 310.70 &  19.52 \\
Meteora     & 19.37 & 32.92 & 0.08 & 0.86 & -17.80 &  60.44 &  98.57 \\
PCGRAD      & 29.51 & 31.80 & 0.08 & 0.83 &  -8.27 &  21.50 &  29.50 \\
CAGRAD      & 32.70 & 33.68 & 0.08 & 0.83 &  -7.40 &  21.50 &  29.50 \\
Nash-MTL    & 31.32 & 32.40 & 0.08 & 0.78 &  -6.40 &  21.50 &  29.50 \\
\midrule
{RobuMTL}  & {34.30} & 28.90 & 0.08 & 0.73 & {+1.30} &  21.50 &  30.93 \\
\textbf{RobuMTL+} & \textbf{37.19} & 27.52 & \textbf{0.07} & \textbf{0.67} & \textbf{+9.70} &  21.50 &  30.93 \\
\bottomrule
\end{tabular}%
}
\end{table}

\subsection{Ablation Study}
To evaluate our approach, we tested on each dataset separately to check its performance against the adverse single task baseline and adverse MTLoRA as shown in Figure \ref{fig:roboMTL_comparison2}. As observed, our approach  maintains its accuracy on clean data without sacrificing performance in favor of robustness, while simultaneously ensuring robustness across most of tasks.
\begin{figure}
    \centering
    \includegraphics[width=\linewidth]{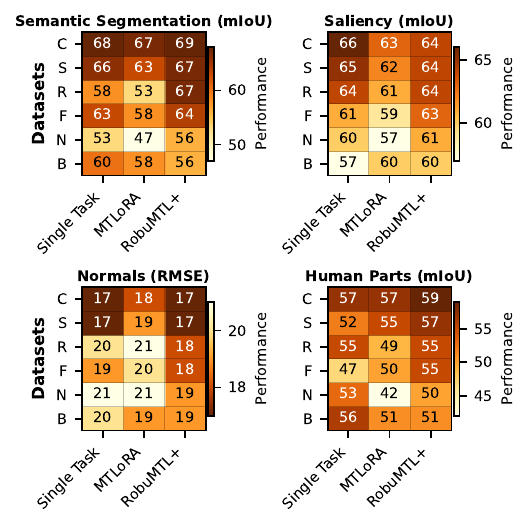}
    \caption{Performance comparison of RobuMTL+ against the baseline and MTLoRA (LoRA rank 16) across multiple datasets. The symbols C, S, R, F, N, B represent the clean, snow, rain, fog, noise, and blur datasets, respectively.}
    \label{fig:roboMTL_comparison2}
\end{figure}
RobuMTL+ outperforms the adverse single-task model by +44.44\% in accuracy under mixed perturbations (rain and fog), as shown in Table \ref{performance_comparison}, using hierarchical ranks (16, 32, 64, and 128). The gain reflects Normals’ sensitivity to mixed perturbations and MTL’s ability especially RobuMTL’s to mitigate them via shared, robust features. {Table $\ref{expertresults}$ shows the performance of aggregating k-experts on the model using hierarchical ranks (8, 16, 32, and 64).} The DMLS assigns high scores to the LoRA expert corresponding to the detected dominant perturbation and low scores to the others, which is why increasing the number of experts ($K$) has only a slight impact on performance. 
\begin{table}
    \centering
      \caption{Performance comparison of RobuMTL+ (with MTLoRA and single-task models under mixed perturbations (PASCAL), including rain and fog.}
    \label{performance_comparison}
    \resizebox{\columnwidth}{!}{  
    \small
        \begin{tabular}{lccccc}
        \hline
        \textbf{Model} & \textbf{Semseg} ↑ & \textbf{Sal} ↑ & \textbf{Norm} ↓ & \textbf{HP} ↑ & \textbf{${\Delta_m}$} \\ \hline
        Single Task & 43.05 & 59.58 & 105.9 & 40.02 & 0 \\ 
        MTLoRA & 40.64 & 57.93 & 23.03 & 42.01 & +16.24  \\ 
        \textbf{RobuMTL+} & \textbf{64.78} & \textbf{64.18} & \textbf{19.05} & \textbf{54.99} & \textbf{+44.40} \\ \hline
        \end{tabular}
    }
  
\end{table}
However, merging all experts reduces performance on single perturbations, where only one expert is relevant, but improves robustness on mixed perturbations through complementary contributions.

\begin{table}[h!]
\centering
\caption{Comparison of $\Delta_{\text{adv}}$ of RobuMTL+ under single and mixed perturbations across $k$ values on PASCAL.} 
\label{expertresults}
\resizebox{\columnwidth}{!}{
\begin{tabular}{c|c|c|c|c|c|c} 
\hline
 & \textbf{$k=1$} & \textbf{$k=2$} & \textbf{$k=3$} & \textbf{$k=4$} & \textbf{$k=5$} & \textbf{$k=6$} \\
\hline
$\Delta_{\text{adv}}$(\textbf{Single}) & +3.10 & +0.82 & +0.80 & +0.72 & +0.70 & -3.10 \\ \hline
$\Delta_{\text{adv}}$(\textbf{Mixed})  & +44.00 & +44.30 & +44.33 & +44.36 & +44.38 & +44.40 \\
\hline
\end{tabular}}
\end{table}

We observed that conventional routers within MoE are unable to reliably differentiate between perturbation patterns, which leads to poor expert selection and significant degradation in performance. In fact, most of the performance drop in standard multi-task MoE models stems from the router’s failure to separate perturbation-specific features. By introducing our specialized router, we improve this expert selection process and achieve up to a 60\% performance boost, as shown in Table \ref{tab:pascal2}.
 \begin{table}[h!]
\centering
\caption{{PASCAL performance of MTLMOE, single task, single task with LoRA after using DMLS+MEPF (ours) based on swinT.} 
}
\label{tab:pascal2}
\resizebox{\columnwidth}{!}{
\begin{tabular}{lccccc}
\toprule
\textbf{Model} & \textbf{Semseg} ↑ & \textbf{Norm} ↓ & \textbf{Sal} ↑ & \textbf{HP} ↑ &\textbf{${\Delta_m}$} \\
\midrule
MTLMOE+ours  & {62} & {19.4} & {74.6} & {53}&$-$1.4   \\
STMOE+ours  &  62.4 &18.9&76.6&52.6&+0.6\\
\bottomrule
\end{tabular}}
\end{table}

 
Table \ref{tab:pascal3} shows the effect of each component on RobuMTL performance. Adding the lightweight DMLS increases robustness, while MEPFM and Squad produce the largest performance gains. Normal expert averaging lowers accuracy by diluting perturbation specific adaptations.
 \begin{table}[h!]
\centering
\caption{{Effect of each component of RobuMTL(+) on overall performance on PASCAL.}
}
\label{tab:pascal3}
\resizebox{\columnwidth}{!}{
\begin{tabular}{lccccc}
\toprule
\textbf{Model}  &\textbf{${\Delta_m}$} & \textbf{Param(M)}& \textbf{Flops(G)}
\\
\midrule
MTL+clean LoRA expert & -12.78 &30.87&20.51   \\
+ rest of experts & -7.25&30.87 &20.51 \\
+ dmls w/o se (+ avg K LoRAs) &-6.0 &30.9& 21.5
\\
+ dmls w/ se (+ avg K LoRAs) &-5.71&30.92& 21.5 
\\
+MEPF&+1.5&30.92& 21.5 
\\
+Squad&+2.8&30.92& 21.5 
\\
\bottomrule
\end{tabular}}
\end{table}
\begin{figure}[h!]
    \centering
    \includegraphics[width=\linewidth]{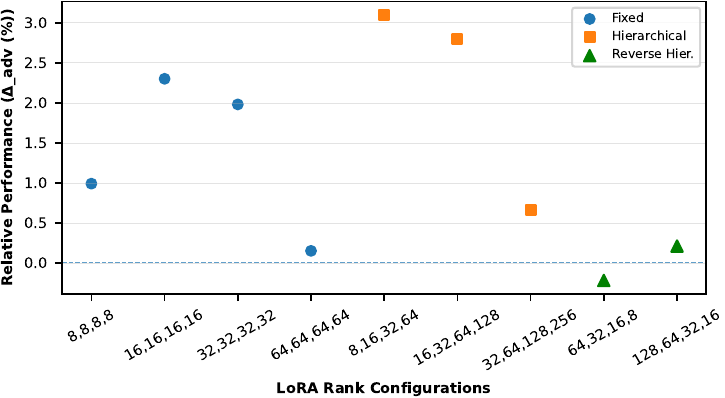}
    \caption{Performance of different ranks on PASCAL.}
    \label{fig:rank_results}
\end{figure} 
Figure \ref{fig:rank_results} illustrates how rank distribution affects model performance. With fixed ranks, results vary, and rank = 16 achieves the strongest overall performance. For hierarchical rank settings, smaller ranks in early layers reduce noise propagation, whereas tasks such as normals and semantic segmentation degrade more severely under snow, noise, and blur when early-layer ranks are too large. Although the rank (8,16,32,64) configuration yields the highest robustness, its clean accuracy is lower than rank (16,32,64,128). Therefore, we select rank (16,32,64,128) in Table \ref{table:results} as the best trade-off between clean and robust performance. The full breakdown supporting this conclusion can be found in the supplementary materials.

\section{Conclusion} \label{conclusion}
In this work, we demonstrated that both multi-task and single-task models significantly degrade under adverse weather conditions. To enhance robustness without sacrificing clean performance, we introduced RobuMTL, a hierarchical LoRA-based adaptive framework that dynamically selects specialized LoRA experts for each input. Our approach achieves up to 2.8\% improvement under adversarial conditions and 13.5\% under clean settings on PASCAL, as well as 9.7\% gains on NYUDv2 and 44.4\% under mixed perturbations on PASCAL. In the future, we plan to extend RobuMTL to real-world weather datasets.

\newpage
\section*{Acknowledgment} \label{ack}
This work is partially supported by NSF grant numbers $2453413$ and $2350180$.

{
    \small
    \bibliographystyle{ieeenat_fullname}
    \bibliography{main}
}
\newpage

\clearpage
\twocolumn[
\begin{center}
    {\large \bfseries RobuMTL: Enhancing Multi-Task Learning Robustness \\ Against Weather Conditions \\[0.8em]}
    {\large \itshape Supplementary Materials}
    \vspace{0.8cm}
\end{center}
]
\subsection*{DMLS Performance}
The performance of the DMLS is summarized in Table \ref{dmls_table}, where we evaluate the model across several key metrics: Accuracy, Precision, Recall, and F1-score. Accuracy reflects the model’s ability to make correct predictions, providing a general measure of its overall effectiveness.
Precision quantifies how many of the instances predicted as positive are actually correct, while recall measures how well the model identifies all true positive instances. The F1-score, calculated as the harmonic mean of precision and recall, balances these two metrics to provide a comprehensive assessment of the model’s performance.
The Receiver Operating Characteristic (ROC) curve illustrates the capability of DMLS in discriminate between perturbation classes as shown in Figure \ref{fig:roc}. The DMLS provides good classification performance on rain, noise, blur, and fog classes. It may struggle slightly to differentiate between snow and clean images, as they share some common properties, and some images contain small snowflake patterns that lead the classifier to mistakenly classify them as clean. However, the 6\% difference in accuracy does not significantly impact the performance of routing to select the top-K experts.  This is because the scores are generated based on the total voting over the batch of images. Even if the classifier misclassifies one of the images, the overall voting across the batch will mitigate this issue achieving 99.9\% accuracy.
\begin{figure}[ht]
\centering 
\captionsetup{justification=centering}
\includegraphics[width=\columnwidth]{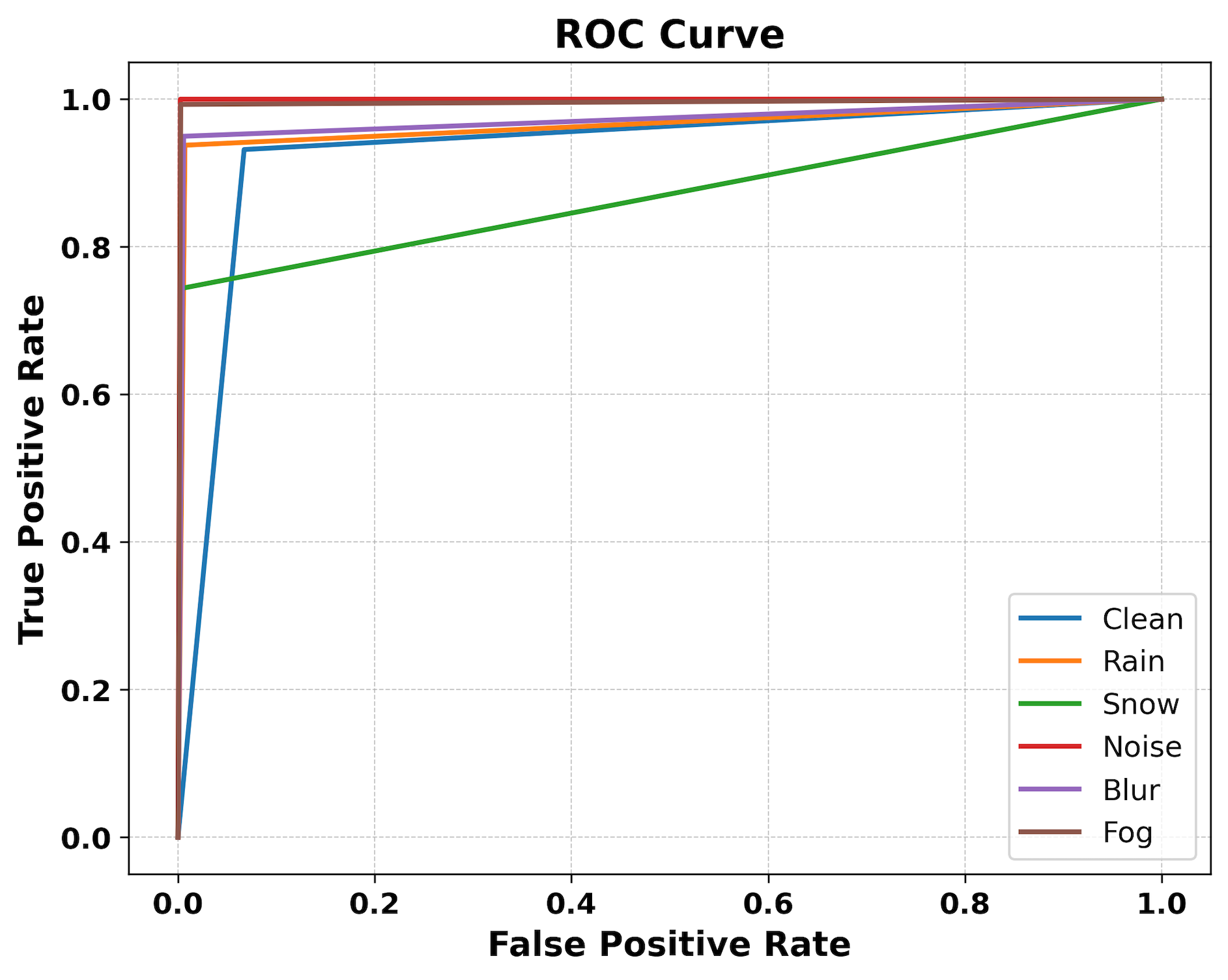} 
\caption{The DMLS performance ROC curve in classification of perturbations.}
\label{fig:roc}
\end{figure}
\begin{table}[h]
   \begin{minipage}{0.4\linewidth}  

    \end{minipage}
    \centering
    \begin{minipage}{0.55\linewidth}  
        \centering
        \caption{DMLS performance metrics.} \label{dmls_table}
        \begin{tabular}{c|c}
            \hline
            \textbf{Metric} & \textbf{Value} \\
            \hline
            Accuracy (\%) & 94.44 \\
            Precision (\%) & 93.36 \\
            Recall (\%) & 92.58 \\
            F1-score (\%) & 92.65 \\
            No. Parameters  & 48,742 \\
            \hline
        \end{tabular}
    \end{minipage}%
    \hfill
 
\end{table}

    


\subsection*{LoRA Ranks Performance on Perturbations}
From the shown Figure \ref{config_ranks}, the LoRA configuration r[64,64,64,64] achieves the best overall performance on the Human Parts segmentation task under clean conditions, indicating that keeping a consistently strong representational capacity across all encoder stages benefits high-level semantic recognition. The r[8,16,32,64] configuration performs better under perturbations, especially noise and blur, but shows a noticeable drop in clean performance due to its lower capacity in early layers. The r[16,32,64,128] setup provides a balanced alternative, ranking third overall, and shows competitive robustness under certain weather degradations, but still does not reach the same clean-data peak as uniform higher ranks.

These trends reveal that Human Parts task relies heavily on global structural context and fine-grained semantic cues, which become more stable when the model maintains sufficient rank in shallow and mid-level stages (as in r[64,64,64,64]). Meanwhile, hierarchical growth schemes like r[8,16,32,64] and r[16,32,64,128] introduce useful regularization against high-frequency perturbations by constraining early representations and allowing more flexibility deeper but this comes at the cost of reduced clean-scene accuracy for a task that demands strong semantic detail at all feature scales.
\begin{figure*}[!h]
    \centering

    \begin{subfigure}{0.47\linewidth}
        \centering
        \includegraphics[width=\linewidth]{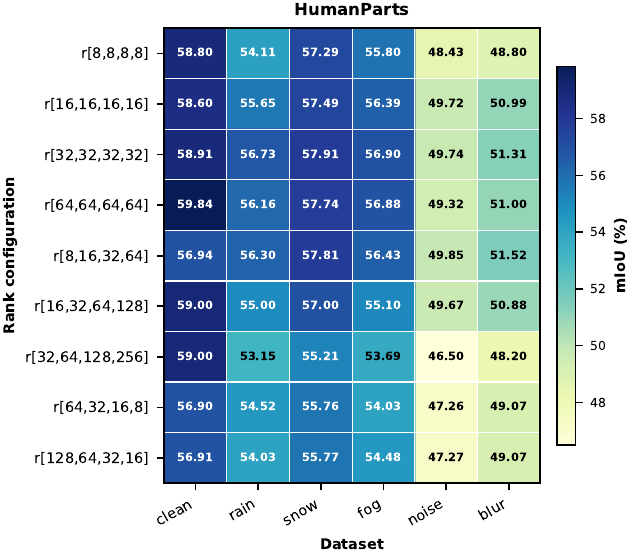}
        \caption{Human Parts task performance on different ranks.} 
    \end{subfigure}
    \hfill
    \begin{subfigure}{0.47\linewidth}
        \centering
        \includegraphics[width=\linewidth]{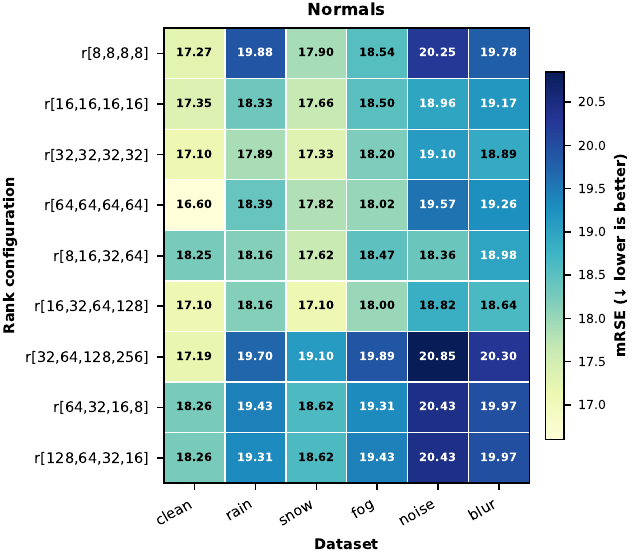}
        \caption{Normals task performance on different ranks.}
    \end{subfigure}

    \begin{subfigure}{0.47\linewidth}
        \centering
        \includegraphics[width=\linewidth]{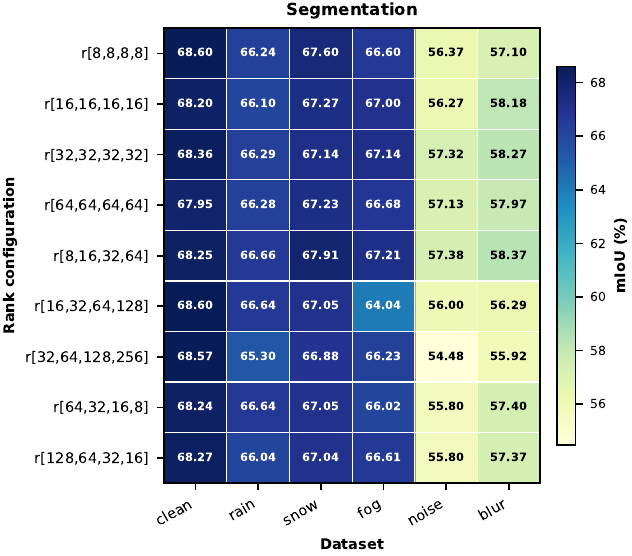}
        \caption{Semantic Segmentation task performance on different ranks.}
    \end{subfigure}
    \hfill
    \begin{subfigure}{0.47\linewidth}
        \centering
        \includegraphics[width=\linewidth]{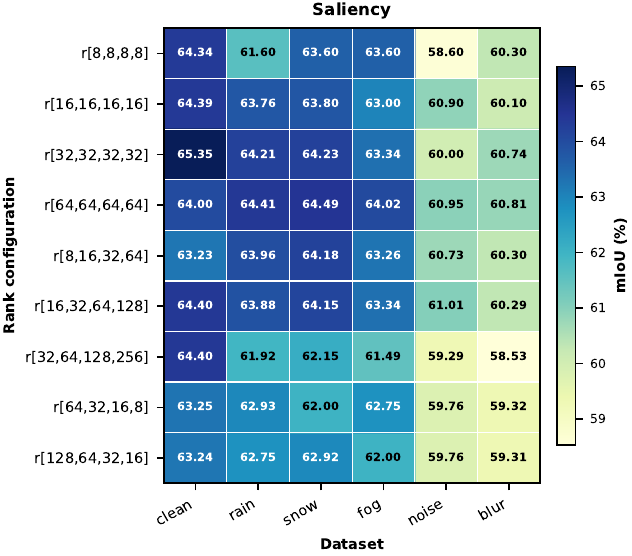}
        \caption{Saliency task performance on different ranks.}
    \end{subfigure}

    \caption{Breakdown of task performance across different LoRA rank configurations on clean and perturbed PASCAL datasets.} \label{config_ranks}
    \label{fig:2x2}
\end{figure*}
For the Normals task, the r[64,64,64,64] configuration produces the lowest error in clean conditions, but it is less robust against perturbations, particularly noise and blur, where its error noticeably increases. In contrast, r[16,32,64,128] provides the best overall trade-off between clean accuracy and robustness, achieving strong performance across multiple adverse conditions including snow, fog, and blur. Meanwhile, the r[8,16,32,64] hierarchy demonstrates high robustness against rain, snow, and noise, benefiting from stronger suppression of high-frequency corruption in the earliest layers. 
These trends reflect the sensitivity of surface normal estimation to low-level geometric consistency. Uniform high ranks allow the network to preserve rich detail under clean conditions, but the same flexibility can lead to over-amplifying noise when early-stage representations are not sufficiently constrained. Hierarchically increasing ranks (as in r[16,32,64,128]) strike a favorable balance by limiting noise propagation in shallow layers while expanding representational capacity deeper in the network where shape and orientation information is consolidated. Additionally, the strong noise robustness of r[8,16,32,64] suggests that aggressive early compression acts like a built-in denoising filter, preventing corrupted features from degrading later geometric predictions.

For the saliency estimation task, the r[32,32,32,32] configuration delivers the highest accuracy under clean conditions and ranks among the best for several perturbations, but it shows a notable drop in robustness when strong noise degradations are introduced. Both r[8,16,32,64] and r[16,32,64,128] follow closely behind, maintaining stronger resilience against most perturbations, especially those that distort structural boundaries. In contrast, rank settings such as r[8,16,32,64], r[64,32,16,8], and r[128,64,32,16] demonstrate lower clean-data accuracy, indicating that either overly limited early-layer capacity or reversed/imbalanced rank hierarchies can weaken saliency recognition in ideal conditions.

This behavior aligns with the nature of saliency detection, which requires both mid-level semantic consistency and preservation of fine spatial cues. A uniform moderate rank like r[32,32,32,32] offers the right amount of global context extraction and detail representation, explaining its superior clean performance. However, the improved robustness of r[8,16,32,64] and r[16,32,64,128] suggests that progressively increasing ranks across the network help buffer against corruption, where early compression filters out noise and deeper layers retain enough flexibility to reconstruct meaningful foreground attention cues.

For the segmentation task, the r[8,16,32,64] configuration achieves the highest robustness across all perturbation types, showing strong accuracy even under challenging degradation, with only a slight decrease in clean performance. The r[8,8,8,8] configuration follows as the next most consistent choice under corruption, demonstrating that a uniformly low-rank design can effectively filter noise, though it lacks the deeper representational strength needed for peak accuracy in clean scenes. Meanwhile, r[16,32,64,128] delivers the best clean-data performance and excels specifically under snow, benefiting from increased rank capacity in deeper layers where semantic boundaries are refined.

This pattern highlights a key insight: segmentation models require a balance between noise suppression in early feature extraction and semantic expressiveness in deeper layers. Increasing ranks hierarchically (as in r[8,16,32,64]) allows the model to localize corruptions early while progressively restoring discriminative features, making it the most reliable choice across perturbations. In contrast, while higher ranks throughout the network (as in r[16,32,64,128]) can maximize clean-scene accuracy, they become more sensitive to corruption due to the higher freedom in feature adaptation, causing performance to deteriorate more noticeably under harsh perturbations.

Overall, our results consistently demonstrate that forward hierarchical LoRA rank configurations outperform uniform and reverse configurations in multi-task learning. In particular, the r[8,16,32,64] setup delivers the highest relative performance gains under perturbations across all tasks, effectively suppressing high-frequency noise in early layers and enabling deeper layers to recover semantic structure, although it exhibits slightly lower accuracy on clean data. Meanwhile, r[16,32,64,128] achieves the best balance between clean-scene performance and robustness, offering both strong baseline accuracy and stable degradation behavior across weather-related corruptions. These findings suggest that progressively increasing rank with network depth is a crucial design principle for enhancing MTL resilience: early-stage compression restricts noise propagation while deeper high-capacity representations preserve task-specific semantic information. In contrast, reverse hierarchical ranks perform consistently worse, indicating that allocating higher representational capacity to shallow layers is neither effective for semantics nor for robustness. Thus, forward hierarchical LoRA assignments, especially r[8,16,32,64] and r[16,32,64,128], provide the most reliable and scalable adaptation strategies for perturbation-aware multi-task learning



\subsection*{Frame Per Second (FPS) Evaluation}

In our evaluation, the Frames Per Second (FPS) metric was calculated based on the average inference time measured over randomly sampled and combined images from all perturbation datasets, following:
\begin{gather*}
\text{FPS} = \frac{NB * BS }{T_{\text{seconds}}},
\end{gather*}
where $NB$ is the number of processed batches, $BS$ is the batch size, and $T_{\text{seconds}}$ is the total time taken to process the frame.
Unlike the original baseline, which reports FPS using batch size = 1, the model gives higher throughput using batched inference as well, as GPUs process multiple images more efficiently in parallel. To ensure fairness in comparison, we did all the evaluation on same GPU and same number CPU cores.
Also, we treat each task-specific single-task model independently and report the minimum FPS among them. Importantly, the MTL model achieves faster overall inference than single-task deployment because it shares the encoder and common computations across tasks, reducing redundant processing. Although our proposed MTL variant achieves slightly lower FPS than the baseline MTL, this overhead is expected due to the additional modules (DMLS and MEFP), which increase robustness and task interaction at a modest cost in speed. RobuMTL(+) adds just 3.6 ms per image (17 ms total), with 0.5 ms for DMLS, 2.5 ms for
LoRA aggregation, and 0.6 ms for injection, enabling robust inference with minimal overhead. 
\subsection*{Model Performance Evaluation}
We evaluate performance by averaging results across each perturbation type for both PASCAL and NYUD-v2. Unlike PASCAL, NYUD-v2 is smaller and includes edge and depth tasks, which make the MTL setup more sensitive to feature degradation. Although the same strategy is applied to both datasets, NYUD-v2 requires slightly higher ranks in the early layers due to its different data distribution and the need for stronger representation in edge and depth tasks.

\subsection*{Model Training}
We explored multiple training paradigms and found that training noise- and blur-specific experts solely on their own data led to overfitting and degraded performance in tasks such as normals. In addition, applying standard MTL conflict-resolution techniques further worsened results, as noise amplification increased the RMSE for normals, depth, and edges.

We also experimented with an auxiliary consistency loss to align the model under perturbations with a clean teacher model. Initial observations showed up to a 1\% improvement in some tasks, but this requires further investigation and analysis as part of future work.

\end{document}